\documentclass[11pt,a4paper]{article}
\usepackage[hyperref]{AACL-IJCNLP2020}
\usepackage{times}
\usepackage{latexsym}
\usepackage{CJKutf8}
\usepackage{multirow}
\usepackage{multicol}
\usepackage{lineno}
\usepackage{datetime}
\usepackage{bm}
\usepackage{makecell}
\usepackage{graphicx}
\newcommand*{\cn}[1]{%
  \begin{CJK}{UTF8}{gbsn}#1\end{CJK}}

\usepackage{microtype}

\aclfinalcopy 
\def\aclpaperid{130} 

\title{Chinese Grammatical Correction Using BERT-based Pre-trained Model}

\author{Hongfei Wang, Michiki Kurosawa\thanks{\quad Currently at Nomura Research Institute, Ltd.}, Satoru Katsumata\thanks{\quad Currently at Retrieva, Inc.} \and Mamoru Komachi \\
  Tokyo Metropolitan University \\
  \texttt{wang-hongfei@ed.tmu.ac.jp},
  \texttt{m-kurosawa@nri.co.jp}\\
  \texttt{satoru.katsumata@retrieva.jp},
  \texttt{komachi@tmu.ac.jp} \\}

\date{}
\begin{document}
\maketitle
\begin{abstract}
In recent years, pre-trained models have been extensively studied, and several downstream tasks have benefited from their utilization. In this study, we verify the effectiveness of two methods that incorporate a BERT-based pre-trained model developed by \citet{cui-etal-2020-revisiting} into an encoder-decoder model on Chinese grammatical error correction tasks. We also analyze the error type and conclude that sentence-level errors are yet to be addressed.
\end{abstract}

\section{Introduction}

Grammatical error correction (GEC) can be regarded as a sequence-to-sequence task. GEC systems receive an erroneous sentence written by a language learner and output the corrected sentence. In previous studies that adopted neural models for Chinese GEC \citep{ren_sequence_2018, Zhou2018ChineseGE}, the performance was improved by initializing the models with a distributed word representation, such as Word2Vec \citep{mikolov2013distributed}. However, in these methods, only the embedding layer of a pre-trained model was used to initialize the models. 

In recent years, pre-trained models based on Bidirectional Encoder Representations from Transformers (BERT) have been studied extensively \citep{Devlin2019BERTPO, Liu2019RoBERTaAR}, and the performance of many downstream Natural Language Processing (NLP) tasks has been dramatically improved by utilizing these pre-trained models. To learn existing knowledge of a language, a BERT-based pre-trained model is trained on a large-scale corpus using the encoder of Transformer \citep{Vaswani2017AttentionIA}. Subsequently, for a downstream task, a neural network model is initialized with the weights learned by a pre-trained model that has the same structure and is fine-tuned on training data of the downstream task. Using this two-stage method, the performance is expected to improve because downstream tasks are informed by the knowledge learned by the pre-trained model.

Recent works \citep{Kaneko2020EncoderDecoderMC, Kantor2019LearningTC} show that BERT helps improve the performance on the English GEC task. As the Chinese pre-trained models are developed and released continuously \citep{cui-etal-2020-revisiting, zhang-etal-2019-ernie}, the Chinese GEC task may also benefit from using those pre-trained models.

\begin{figure}[t]
    \centering
    \includegraphics[width=7.7cm]{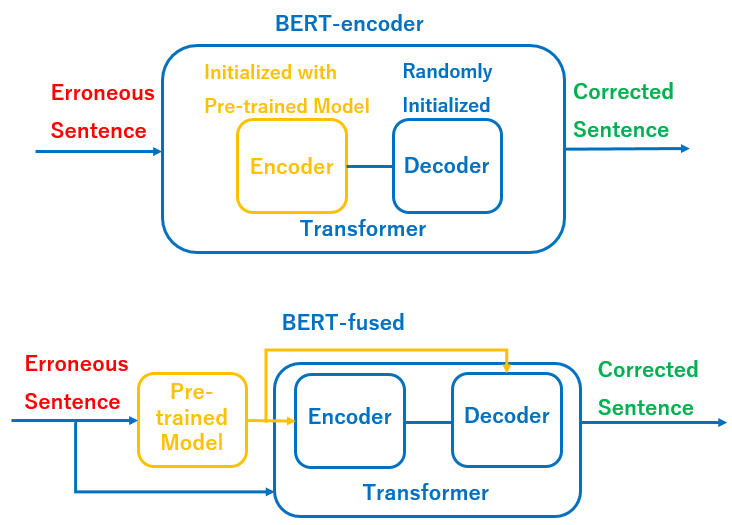}
    \caption{Two methods for incorporating a pre-trained model into the GEC model.}
    \label{fig: model}
\end{figure}{}

In this study, as shown in Figure \ref{fig: model}, we develop a Chinese GEC model based on Transformer with a pre-trained model using two methods: first, by initializing the encoder with the pre-trained model (BERT-encoder); second, by utilizing the technique proposed by \citet{Zhu2020Incorporating}, which uses the pre-trained model for additional features (BERT-fused); on the  Natural Language Processing and Chinese Computing (NLPCC) 2018 Grammatical Error Correction shared task test dataset \citep{Zhao2018OverviewOT}, our single models obtain $\mathrm{F_{0.5}}$ scores of 29.76 and 29.94 respectively, which is similar to the performance of ensemble models developed by the top team of the shared task. Moreover, using a 4-ensemble model, we obtain an $\mathrm{F_{0.5}}$ score of 35.51, which outperforms the results from the top team  by a large margin. We annotate the error types of the development data; the results show that word-level errors dominate all error types and that sentence-level errors remain challenging and require a stronger approach.

\section{Related Work}

Given the success of the shared tasks on English GEC at the Conference on Natural Language Learning (CoNLL) \citep{ng-etal-2013-conll,ng-etal-2014-conll}, a Chinese GEC shared task was performed at the NLPCC 2018. In this task, approximately one million sentences from the language learning website Lang-8\footnote{\url{https://lang-8.com/}} were used as training data and two thousand sentences from the PKU Chinese Learner Corpus \citep{Zhao2018OverviewOT} were used as test data. Here, we briefly describe the three methods with the highest performance.

First, \citet{Fu2018YoudaosWS} combined a 5-gram language model-based spell checker with subword-level and character-level encoder-decoder models using Transformer to obtain five types of outputs. Then, they re-ranked these outputs using the language model. Although they reported a high performance, several models were required, and the combination method was complex.

Second, \citet{ren_sequence_2018} utilized a convolutional neural network (CNN), such as in \citet{chollampatt2018multilayer}. However, because the structure of the CNN is different from that of BERT, it cannot be initialized with the weights learned by the BERT.

Last, \citet{Zhao2020MaskGECIN} proposed a dynamic masking method that replaces the tokens in the source sentences of the training data with other tokens (e.g. [PAD] token). They achieved state-of-the-art results on the NLPCC 2018 Grammar Error Correction shared task without using any extra knowledge. This is a data augmentation method that can be a supplement for our study.

\section{Methods}
In the proposed method, we construct a correction model using Transformer, and incorporate a Chinese pre-trained model developed by \citet{cui-etal-2020-revisiting} in two ways as described in the following sections.
\subsection{Chinese Pre-trained Model}

We use a BERT-based model as our pre-trained model. BERT is mainly trained with a task called Masked Language Model. In the Masked Language Model task, some tokens in a sentence are replaced with masked tokens ([MASK]) and the model has to predict the replaced tokens. 

In this study, we use the Chinese-RoBERTa-wwm-ext model developed by \citet{cui-etal-2020-revisiting}. The main difference between Chinese-RoBERTa-wwm-ext and the original BERT is that the latter uses whole word masking (WWM) to train the model. In WWM, when a Chinese character is masked, other Chinese characters that belong to the same word should also be masked.

\subsection{Grammatical Error Correction Model}
\label{our-model}
In this study, we use Transformer as the correction model. Transformer has shown excellent performance in sequence-to-sequence tasks, such as machine translation, and has been widely adopted in recent studies on English GEC \citep{kiyono-etal-2019-empirical,junczys-dowmunt-etal-2018-approaching}.

However, a BERT-based pre-trained model only uses the encoder of Transformer; therefore, it cannot be directly applied to sequence-to-sequence tasks that require both an encoder and a decoder, such as GEC. Hence, we incorporate the encoder-decoder model with the pre-trained model in two ways as described in the following subsections.

\paragraph{BERT-encoder}
We initialize the encoder of Transformer with the parameters learned by Chinese-RoBERTa-wwm-ext; the decoder is initialized randomly. Finally, we fine-tune the initialized model on Chinese GEC data.

\paragraph{BERT-fused}
\citet{Zhu2020Incorporating} proposed a method that uses a pre-trained model as the additional features. In this method, input sentences are fed into the pre-trained model and representations from the last layer of the pre-trained model are acquired first. Then, the representations will interact with the encoder and decoder by using attention mechanism. \citet{Kaneko2020EncoderDecoderMC} verified the effectiveness of this method on English GEC tasks. 

\section{Experiments}
\subsection{Experimental Settings}
\paragraph{Data}
In this study, we use the data provided by the NLPCC 2018 Grammatical Error Correction shared task. We first segment all sentences into characters because the Chinese pre-trained model we used is character-based. In the GEC task, source and target sentences do not tend to change significantly. Considering this, we filter the training data by excluding sentence pairs that meet the following criteria: i) the source sentence is identical to the target sentence; ii) the edit distance between the source sentence and the target sentence is greater than 15; iii) the number of characters of the source sentence or the target sentence exceeds 64. Once the training data were filtered, we obtained 971,318 sentence pairs. 

Because the NLPCC 2018 Grammatical Error Correction shared task did not provide development data, we opted to randomly extract 5,000 sentences from the training data as the development data following \citet{ren_sequence_2018}.

The test data consist of 2,000 sentences extracted from the PKU Chinese Learner Corpus. According to \citet{Zhao2018OverviewOT}, the annotation guidelines follow the minimum edit distance principle \citep{nagata-sakaguchi-2016-phrase}, which selects the edit operation that minimizes the edit distance from the original sentence.

\paragraph{Model}
We implement the Transformer model using fairseq 0.8.0.\footnote{\url{https://github.com/pytorch/fairseq}} and load the pre-trained model using pytorch\_transformer 2.2.0.\footnote{\url{https://github.com/huggingface/transformers}} 

We then train the following models based on Transformer.

\textbf{Baseline}: a plain Transformer model that is initialized randomly without using a pre-trained model.

\textbf{BERT-encoder}: the correction model introduced in Section \ref{our-model}.

\textbf{BERT-fused}: the correction model introduced in Section \ref{our-model}. We use the implementation provided by \citet{Zhu2020Incorporating}.\footnote{\url{https://github.com/bert-nmt/bert-nmt}} 

Finally, we train a 4-ensemble BERT-encoder model and a 4-ensemble BERT-fused model. 

More details on the training are provided in the appendix \ref{sec:appendix}.

\paragraph{Evaluation}
As the evaluation is performed on word-unit, we strip all delimiters from the system output sentences and segment the sentences using the pkunlp\footnote{ \url{http://59.108.48.12/lcwm/pkunlp/downloads/libgrass- ui.tar.gz}} provided in the NLPCC 2018 Grammatical Error Correction shared task.

Based on the setup of the NLPCC 2018 Grammatical Error Correction shared task, the evaluation is conducted using \textit{MaxMatch} (M2).\footnote{\url{https://github.com/nusnlp/m2scorer}}

\subsection{Evaluation Results}

Table \ref{table:results} summarizes the experimental results of our models. We run the single models four times, and report the average score. For comparison, we also cite the result of the state-of-the-art model \citep{Zhao2020MaskGECIN} and the results of the models developed by two teams in the NLPCC 2018 Grammatical Error Correction shared task.

\begin{table}[t]
\small
\begin{tabular}{l|c|c|c}
\hline
 \textbf{[Our models]}&\textbf{P}&\textbf{R}&\bm{$\mathrm{F_{0.5}}$}\\
Baseline & 25.14 & 14.34 & 21.85 \\
BERT-encoder & 32.67 & 22.19 & 29.76 \\
BERT-fused & 32.11 & \textbf{23.57} & 29.94 \\
BERT-encoder (4-ensemble) & 41.94 & 22.02 & 35.51 \\ 
BERT-fused (4-ensemble) & 32.20 & 23.16 & 29.87 \\ 
\hline
\textbf{[SOTA Result]}\\
\citet{Zhao2020MaskGECIN} & 44.36 & 22.18 & \textbf{36.97}\\ \hline
\textbf{[NLPCC 2018]}\\
\citet{Fu2018YoudaosWS} & 35.24 & 18.64 & 29.91\\
\citet{ren_sequence_2018} & 41.73 & 13.08 & 29.02 \\
\citet{ren_sequence_2018} (4-ensemble)& \textbf{47.63} & 12.56 & 30.57\\\hline
\end{tabular}
\caption{Experimental results on the NLPCC 2018 Grammatical Error Correction shared task.}
\label{table:results}
\end{table}

\begin{table*}[t]
 \centering
    \small
    \begin{tabular}{c|c|c}
         \hline
         src & \cn{\textbf{持 别} 是 北京 ， 没有 “ 自然 ” 的 感觉 。} & \cn{人们 在 一 辈子 \textbf{经验} 很多 事情 。}\\
         gold& \cn{\textbf{特别} 是 北京 ， 没有 “ 自然 ” 的 感觉 。} &  \cn{人们 在 一 辈子 \textbf{经历} 很多 事情 。}\\    
         baseline&\cn{\textbf{持 别} 是 北京 ， 没有 “ 自然 ” 的 感觉 。} & \cn{人们 在 一辈子 \textbf{经历} 了 很多 事情 。}\\ 
         BERT-encoder&\cn{\textbf{特别} 是 北京 ， 没有 “ 自然 ” 的 感觉 。} & \cn{人们 一辈子 \textbf{会} \textbf{经历} 很多 事情 。}\\
         Translation&\textbf{Especially} in Beijing, there is no \textit{natural} feeling. & People \textbf{experience} many things in their lifetime.\\
         \hline

\end{tabular}
\caption{Source sentence, gold edit, and output of our models.}
\label{table:example}
\end{table*}

\begin{table*}
\small
\centering
\begin{tabular}{c|c|l}
     \hline
     \textbf{Error Type} & \makecell[c]{\textbf{Number of} \\\textbf{errors}} & \textbf{Examples} \\\hline
     B & 9 & \makecell[l]{\cn{最后 ， 要 \underline{关主}\{\textbf{关注}\} 一些 关于 天气 预报 的 新闻 。}\\ (Finally, pay attention to some weather forecast news.)}\\\hline
     
     CC & 35 & \makecell[l]{\cn{有 一 天 晚上 他 下 了 \underline{决定}\{\textbf{决心}\} 向 富丽 堂皇 的 宫殿 里 走 ， 偷偷 \underline{的}\{\textbf{地}\} 进入}\\\cn{宫内 。} (One night he decided to walk to the magnificent palace, and sneaked in it secretly.)}\\\hline
     
     CQ & 30 & \makecell[l]{\cn{在 上海 我 总是 住 \underline{NONE}\{\textbf{在}\} 一家 特定 \underline{NONE}\{\textbf{的}\} 酒店 。} \\(I always stay in the same hotel in Shanghai.)}
     \\\hline
     CD & 21 & \makecell[l]{\cn{我 很 喜欢 \underline{念}\{\textbf{NONE}\}读 小说 .} (I like to read novels.)} \\\hline
     CJ & 35 & \makecell[l]{\cn{…… 但是 同时 也 对 环境 \underline{问题}\{\textbf{NONE}\} \underline{日益 严重 造成 了}\{\textbf{造成 了 日益 严重 的}\}}\\ \cn{空气 污染 问题 。} (But on the meanwhile, it also aggravated the problem of air pollution.)} \\\hline

\end{tabular}{}
\caption{Examples of each error type. The underlined tokens are detected errors that should be replaced with the tokens in braces.}
\label{table: example}
\end{table*}

\begin{table}
\small
\centering
\begin{tabular}{c|c|c|c|c|c|c}
\hline
\multirow{2}{*}{\textbf{Type}}  & \multicolumn{3}{c|}{\textbf{Detection}} & \multicolumn{3}{c}{\textbf{Correction}} \\
\cline{2-4} \cline{5-7}
& \textbf{P} & \textbf{R} & \bm{$\mathrm{F_{0.5}}$}& \textbf{P} & \textbf{R} & \bm{$\mathrm{F_{0.5}}$} \\\hline
\multicolumn{7}{l}{\textbf{BERT-encoder}}\\\hline
B & \textbf{80.0} & \textbf{55.6} & \textbf{73.5} & \textbf{80.0} & \textbf{55.6} & \textbf{73.5}  \\
CC & 62.5 & 31.4 & 52.2 & 43.8 & 20.0 & 35.4 \\
CQ & 65.0 & 43.3 & 59.1 & 45.0 & 30.0 & 40.9 \\
CD & 58.3 & 28.6 & 48.3 & 50.0 & 28.6 & 43.5\\
CJ & 56.5 & 42.9 & 53.1 & \makecell[r]{4.3} & \makecell[r]{2.9}& \makecell[r]{3.9}\\
\hline
\multicolumn{7}{l}{\textbf{BERT-fused}}\\\hline
B & \textbf{80.0} & 44.4 & \textbf{69.0} & \textbf{80.0} & 44.4 & \textbf{69.0}  \\
CC & 61.9 & 42.9 & 56.9 & 38.1 & 22.9 & 33.6 \\
CQ & 69.0 & \textbf{63.3} & 67.8 & 44.8 & \textbf{46.7} & 45.2 \\
CD & 71.4 & 42.9 & 63.0 & 57.1 & 38.1 & 51.9\\
CJ & 63.2 & 34.3 & 54.1 & 15.8 & \makecell[r]{8.6} & 13.5\\
\hline
\end{tabular}
\caption{Detection and correction performance of BERT-encoder and BERT-fused models on each type of error. }
\label{table:error type}
\end{table}

The performances of BERT-encoder and BERT-fused are significantly superior to that of the baseline model and are comparable to those achieved by the two teams in the NLPCC 2018 Grammatical Error Correction shared task, indicating the effectiveness of adopting the pre-trained model. 

The BERT-encoder (4-ensemble) model yields an $\mathrm{F_ {0.5}}$ score nearly 5 points higher than the highest-performance model in the NLPCC 2018 Grammatical Error Correction shared task. However, there is no improvement for the BERT-fused (4-ensemble) model compared with the single BERT-fused model. We find that the performance of the BERT-fused model depends on the warm-up model. Compared with \citet{Kaneko2020EncoderDecoderMC} using a state-of-the-art model to warm-up their BERT-fused model, we did not use a warm-up model in this work. The performance noticeably drops when we try to warm-up the BERT-fused model from a weak baseline model, therefore, the BERT-fused model may perform better when warmed-up from a stronger model (e.g., the model proposed by \citet{Zhao2020MaskGECIN}).

For the state-of-the-art result achieved by \citet{Zhao2020MaskGECIN}, both the precision and the recall are comparatively high, and they therefore obtain the best $\mathrm{F_{0.5}}$ score.

Additionally, the precision of the models that used a pre-trained model is lower than that of the models proposed by the two teams; conversely, the recall is significantly higher.

\section{Discussion}
\paragraph{Case Analysis}
Table \ref{table:example} shows the sample outputs.

In the first example, the spelling error \cn{持别} is accurately corrected to \cn{特别} (which means \textit{especially}) by the proposed model, whereas it is not corrected by the baseline model. Hence, it appears that the proposed model captures context more efficiently by using the pre-trained model through the WWM strategy.

In the second example, the output of the proposed model is more fluent, although the correction made by the proposed model is different from the gold edit. The proposed model not only changed the wrong word \cn{经验} (which usually means the noun \textit{experience}) to \cn{经历} (which usually means the verb \textit{experience}), but also added a new word \cn{会} (\textit{would, could}); this addition makes the sentence more fluent. It appears that the proposed model can implement additional changes to the source sentence because the pre-trained model is trained with a large-scale corpus. However, this type of change may affect the precision because the gold edit in this dataset followed the principle of minimum edit distance \citep{Zhao2018OverviewOT}.

\paragraph{Error Type Analysis}
To understand the error distribution of Chinese GEC, we annotate 100 sentences of development data and obtain 130 errors (one sentence may contain more than one error). We refer to the annotation of the HSK learner corpus\footnote{\url{http://hsk.blcu.edu.cn/}} and adopt five categories of error: B, CC, CQ, CD, and CJ. B denotes character-level errors, which are mainly spelling and punctuation errors. CC, CQ, and CD are word-level errors, which are word selection, missed word, and redundant word errors, respectively. CJ denotes sentence-level errors which contain several complex errors, such as word order and lack of subject errors. Several examples are presented in Table \ref{table: example}. Based on the number of errors, it is evident that word-level errors (CC, CQ, and CD) are the most frequent.

 Table \ref{table:error type} lists the detection and correction results of the BERT-encoder and BERT-fused models for each error type. The two models perform poorly on sentence-level errors (CJ), which often involve sentence reconstructions, demonstrating that this is a difficult task. For character-level errors (B), the models achieve better performance than for other error types. Compared with the correction performance, the systems indicate moderate detection performance, demonstrating that the systems address error positions appropriately. With respect to the difference in performance of the two systems on each error type, we can conclude that BERT-encoder performs better on character-level errors (B), and BERT-fused performs better on other error types.

\section{Conclusion}
In this study, we incorporated a pre-trained model into an encoder-decoder model using two methods on Chinese GEC tasks. The experimental results demonstrate the usefulness of the BERT-based pre-trained model in the Chinese GEC task. Additionally, our error type analysis showed that sentence-level errors remain to be addressed.

\section*{Acknowledgments}
This work has been partly supported by the programs of the
Grant-in-Aid for Scientific Research from the Japan Society for the Promotion of Science (JSPS KAKENHI) Grant Numbers 19K12099 and 19KK0286.

\bibliography{aacl-ijcnlp2020}

\begin{thebibliography}{20}
\expandafter\ifx\csname natexlab\endcsname\relax\def\natexlab#1{#1}\fi

\bibitem[{Chollampatt and Ng(2018)}]{chollampatt2018multilayer}
Shamil Chollampatt and Hwee~Tou Ng. 2018.
\newblock A multilayer convolutional encoder-decoder neural network for
  grammatical error correction.
\newblock In \emph{AAAI}.

\bibitem[{Cui et~al.(2020)Cui, Che, Liu, Qin, Wang, and
  Hu}]{cui-etal-2020-revisiting}
Yiming Cui, Wanxiang Che, Ting Liu, Bing Qin, Shijin Wang, and Guoping Hu.
  2020.
\newblock Revisiting pre-trained models for {Chinese} natural language
  processing.
\newblock In \emph{Findings of EMNLP}.

\bibitem[{Devlin et~al.(2019)Devlin, Chang, Lee, and
  Toutanova}]{Devlin2019BERTPO}
Jacob Devlin, Ming-Wei Chang, Kenton Lee, and Kristina Toutanova. 2019.
\newblock {BERT}: Pre-training of deep bidirectional transformers for language
  understanding.
\newblock In \emph{NAACL-HLT}.

\bibitem[{Fu et~al.(2018)Fu, Huang, and Duan}]{Fu2018YoudaosWS}
Kai Fu, Jun Huang, and Yitao Duan. 2018.
\newblock Youdao's winning solution to the {NLPCC}-2018 task 2 challenge: A
  neural machine translation approach to {Chinese} grammatical error
  correction.
\newblock In \emph{NLPCC}.

\bibitem[{Junczys-Dowmunt et~al.(2018)Junczys-Dowmunt, Grundkiewicz, Guha, and
  Heafield}]{junczys-dowmunt-etal-2018-approaching}
Marcin Junczys-Dowmunt, Roman Grundkiewicz, Shubha Guha, and Kenneth Heafield.
  2018.
\newblock Approaching neural grammatical error correction as a low-resource
  machine translation task.
\newblock In \emph{NAACL-HLT}.

\bibitem[{Kaneko et~al.(2020)Kaneko, Mita, Kiyono, Suzuki, and
  Inui}]{Kaneko2020EncoderDecoderMC}
Masahiro Kaneko, Masato Mita, Shun Kiyono, Jun Suzuki, and Kentaro Inui. 2020.
\newblock Encoder-decoder models can benefit from pre-trained masked language
  models in grammatical error correction.
\newblock In \emph{ACL}.

\bibitem[{Kantor et~al.(2019)Kantor, Katz, Choshen, Cohen-Karlik, Liberman,
  Toledo, Menczel, and Slonim}]{Kantor2019LearningTC}
Yoav Kantor, Yoav Katz, Leshem Choshen, Edo Cohen-Karlik, Naftali Liberman,
  Assaf Toledo, Amir Menczel, and Noam Slonim. 2019.
\newblock Learning to combine grammatical error corrections.
\newblock In \emph{BEA@ACL}.

\bibitem[{Kiyono et~al.(2019)Kiyono, Suzuki, Mita, Mizumoto, and
  Inui}]{kiyono-etal-2019-empirical}
Shun Kiyono, Jun Suzuki, Masato Mita, Tomoya Mizumoto, and Kentaro Inui. 2019.
\newblock An empirical study of incorporating pseudo data into grammatical
  error correction.
\newblock In \emph{EMNLP-IJCNLP}.

\bibitem[{Liu et~al.(2019)Liu, Ott, Goyal, Du, Joshi, Chen, Levy, Lewis,
  Zettlemoyer, and Stoyanov}]{Liu2019RoBERTaAR}
Yinhan Liu, Myle Ott, Naman Goyal, Jingfei Du, Mandar Joshi, Danqi Chen, Omer
  Levy, Mike Lewis, Luke~S. Zettlemoyer, and Veselin Stoyanov. 2019.
\newblock {RoBERTa}: A robustly optimized {BERT} pretraining approach.
\newblock \emph{ArXiv}.

\bibitem[{Mikolov et~al.(2013)Mikolov, Sutskever, Chen, Corrado, and
  Dean}]{mikolov2013distributed}
Tomas Mikolov, Ilya Sutskever, Kai Chen, Greg~S Corrado, and Jeff Dean. 2013.
\newblock Distributed representations of words and phrases and their
  compositionality.
\newblock In \emph{NIPS}.

\bibitem[{Nagata and Sakaguchi(2016)}]{nagata-sakaguchi-2016-phrase}
Ryo Nagata and Keisuke Sakaguchi. 2016.
\newblock Phrase structure annotation and parsing for learner {E}nglish.
\newblock In \emph{ACL}.

\bibitem[{Ng et~al.(2014)Ng, Wu, Briscoe, Hadiwinoto, Susanto, and
  Bryant}]{ng-etal-2014-conll}
Hwee~Tou Ng, Siew~Mei Wu, Ted Briscoe, Christian Hadiwinoto, Raymond~Hendy
  Susanto, and Christopher Bryant. 2014.
\newblock The {C}o{NLL}-2014 shared task on grammatical error correction.
\newblock In \emph{CoNLL}.

\bibitem[{Ng et~al.(2013)Ng, Wu, Wu, Hadiwinoto, and
  Tetreault}]{ng-etal-2013-conll}
Hwee~Tou Ng, Siew~Mei Wu, Yuanbin Wu, Christian Hadiwinoto, and Joel Tetreault.
  2013.
\newblock The {C}o{NLL}-2013 shared task on grammatical error correction.
\newblock In \emph{CoNLL}.

\bibitem[{Ren et~al.(2018)Ren, Yang, and Xun}]{ren_sequence_2018}
Hongkai Ren, Liner Yang, and Endong Xun. 2018.
\newblock A sequence to sequence learning for {Chinese} grammatical error
  correction.
\newblock In \emph{NLPCC}.

\bibitem[{Vaswani et~al.(2017)Vaswani, Shazeer, Parmar, Uszkoreit, Jones,
  Gomez, Kaiser, and Polosukhin}]{Vaswani2017AttentionIA}
Ashish Vaswani, Noam Shazeer, Niki Parmar, Jakob Uszkoreit, Llion Jones,
  Aidan~N. Gomez, Lukasz Kaiser, and Illia Polosukhin. 2017.
\newblock Attention is all you need.
\newblock In \emph{NIPS}.

\bibitem[{Zhang et~al.(2019)Zhang, Han, Liu, Jiang, Sun, and
  Liu}]{zhang-etal-2019-ernie}
Zhengyan Zhang, Xu~Han, Zhiyuan Liu, Xin Jiang, Maosong Sun, and Qun Liu. 2019.
\newblock {ERNIE}: {E}nhanced language representation with informative
  entities.
\newblock In \emph{ACL}.

\bibitem[{Zhao et~al.(2018)Zhao, Jiang, Sun, and Wan}]{Zhao2018OverviewOT}
Yuanyuan Zhao, Nan Jiang, Weiwei Sun, and Xiaojun Wan. 2018.
\newblock Overview of the {NLPCC} 2018 shared task: Grammatical error
  correction.
\newblock In \emph{NLPCC}.

\bibitem[{Zhao and Wang(2020)}]{Zhao2020MaskGECIN}
Zewei Zhao and Houfeng Wang. 2020.
\newblock Mask{GEC}: Improving neural grammatical error correction via dynamic
  masking.
\newblock In \emph{AAAI}.

\bibitem[{Zhou et~al.(2018)Zhou, Li, Liu, Bao, Xu, and Li}]{Zhou2018ChineseGE}
Junpei Zhou, Chen Li, Hengyou Liu, Zuyi Bao, Guangwei Xu, and Linlin Li. 2018.
\newblock Chinese grammatical error correction using statistical and neural
  models.
\newblock In \emph{NLPCC}.

\bibitem[{Zhu et~al.(2020)Zhu, Xia, Wu, He, Qin, Zhou, Li, and
  Liu}]{Zhu2020Incorporating}
Jinhua Zhu, Yingce Xia, Lijun Wu, Di~He, Tao Qin, Wengang Zhou, Houqiang Li,
  and Tieyan Liu. 2020.
\newblock Incorporating {BERT} into neural machine translation.
\newblock In \emph{ICLR}.

\end{thebibliography}
\bibliographystyle{acl_natbib}

\appendix
\section{Appendices}
\label{sec:appendix}
Table \ref{table:training} shows the training details for each model.

\begin{table}[h]
\small
\centering
\begin{tabular}{l|l}
\hline
\multicolumn{2}{l}{\textbf{Baseline}} \\\hline
Architecture & Encoder (12-layer), Decoder (12-layer)\\
Learning rate & $1\times10^{-5}$ \\
Batch size & 32\\
Optimizer & Adam ($\beta_1=0.9, \beta_2=0.999, \epsilon=1\times10^{-8}$)\\
Max epochs & 20\\
Loss function & cross-entropy\\
Dropout & 0.1\\
\hline
\multicolumn{2}{l}{\textbf{BERT-encoder}} \\\hline
Architecture & Encoder (12-layer), Decoder (12-layer)\\
Learning rate & $3\times10^{-5}$ \\
Batch size & 32\\
Optimizer & Adam ($\beta_1=0.9, \beta_2=0.999, \epsilon=1\times10^{-8}$)\\
Max epochs & 20\\
Loss function & cross-entropy\\
Dropout & 0.1\\
\hline
\multicolumn{2}{l}{\textbf{BERT-fused}} \\\hline
Architecture & Transformer (big)\\
Learning rate & $3\times10^{-5}$\\
Batch size & 32\\
Optimizer & Adam ($\beta_1=0.9, \beta_2=0.98, \epsilon=1\times10^{-8}$)\\
Max epochs & 20\\
Loss function & label smoothed cross-entropy ($\epsilon_{ls}=0.1$)\\
Dropout & 0.3\\\hline
\end{tabular}
\caption{Training details for each model.}
\label{table:training}
\end{table}

\end{document}